# Deep ensembles based on Stochastic Activation Selection for Polyp Segmentation

Alessandra Lumini[1], Loris Nanni[2]* and Gianluca Maguolo[2]

[1]DISI, Università di Bologna, Via dell'università 50, 47521 Cesena, Italy; alessandra.lumini@unibo.it

[2]DEI, University of Padua, viale Gradenigo 6, Padua, Italy

*Correspondence: loris.nanni@unipd.it;

## ABSTRACT

Semantic segmentation has a wide array of applications ranging from medical-image analysis, scene understanding, autonomous driving and robotic navigation. This work deals with medical image segmentation and in particular with accurate polyp detection and segmentation during colonoscopy examinations. Several convolutional neural network architectures have been proposed to effectively deal with this task and with the problem of segmenting objects at different scale input. The basic architecture in image segmentation consists of an encoder and a decoder: the first uses convolutional filters to extract features from the image, the second is responsible for generating the final output. In this work, we compare some variant of the DeepLab architecture obtained by varying the decoder backbone. We compare several decoder architectures, including ResNet, Xception, EffientNet, MobileNet and we perturb their layers by substituting ReLU activation layers with other functions. The resulting methods are used to create deep ensembles which are shown to be very effective. Our experimental evaluations show that our best ensemble produces good segmentation results by achieving high evaluation scores with a dice coefficient of 0.884, and a mean Intersection over Union (mIoU) of 0.818 for the Kvasir-SEG dataset. To improve reproducibility and research efficiency the MATLAB source code used for this research is available at GitHub: https://github.com/LorisNanni.

## 1. Introduction

Colorectal Cancer is one of the prominent causes of cancer related deaths worldwide. To reduce its occurrence, it is important prevention by screening tests and removal of preneoplastic lesions [1]. Polyps are predecessors to this type of cancers and therefore important to find and accurately segment them through colonoscopy examinations. However, accurate polyp segmentation is a challenging task, due to the large variety of polyps, which may have different shapes, sizes, colors, and textures and appearances. Medical research has classified polyp images in four main classes: adenoma, serrated, hyperplastic, and mixed (rare), but they present are high interclass similarity and intra-class variation. Moreover, the boundary between a polyp and its surrounding mucosa is usually not sharp and there can also be high background object similarity, for instance, where parts of a polyp is covered.

To address these challenges, we propose a semantic image segmentation approach based on deep ensembles of Convolutional Neural Networks (CNN) architectures. Semantic image segmentation refers to the process of associating each individual pixel of an image with a predefined class label. Among several variants semantic segmentation models (i.e. U-Net, Fast Fully-connected network, Gated-SCNN, SegNet and DeepLab) we select DeepLabv3 as base architecture for design different classifiers based on model perturbation.

The objectives of this paper include: (1) employing some of the most popular deep CNN architectures extensively used in computer vision community for semantic image segmentation on medical images; we compare several architectures, including ResNet, Xception, EffientNet, MobileNet; (2) modifying base models by substituting ReLU activation layers with other functions; (3) investigating the feasibility of design deep ensembles by fusing perturbed models; (4) identifying a deep ensemble representing, among others, a high performance approach which can be effectively used polyp segmentation.

The remainder of this paper is organized as follows. Section 2 reviews the most popular deep learning architectures in the computer vision community for semantic image segmentation Section 3 presents our approach for perturbing models and explain of deep ensembles are created; Sections 4 introduces the dataset used in our experiments along with data pre-processing, testing protocols and performance indicators. Sections 5 reports and discusses experimental results. Lastly, Section 6 provides conclusions and future work perspectives.

## 2. Deep Learning for Semantic Image Segmentation

Image segmentation is a pixel-based classification problem which performs pixel-level labeling with a set of object categories for all image pixels. Fully Convolutional Networks (FNC) [2] are one of the first attempts to use CNN for segmentation: they were designed by replacing the last fully connected layers of a net with a fully-convolutional layer that allow the classification of the image on a per-pixel basis.

A step forward in the design of segmentation network is done by the encoder-decoder architecture [3] which overcomes the loss of information of FNC due to the absence of deconvolution, by proposing an architecture where a multi-layer deconvolution network is learned. A similar architecture is proposed by U-Net, a U-shape network where the decoder part downsamples the image and increases the number of features, while the opposite encoder part increases the image resolution to the input size [4]. Another encoder-decoder structure is proposed in SegNet [5], which uses VGG [6] as backbone encoder, coupled to a symmetric decoder structure. In SegNet decoding is performed using max pooling indices from the corresponding encoder layer, as opposed to concatenating it as in U-Net, thus saving memory and getting a better boundary reconstruction.

The next step to image segmentation is represented by DeepLab [7], a semantic segmentation model designed by Google which achieves dense prediction by simply up-sampling the output of the last convolution layer and computing pixel-wise loss. The novelty is in the use of atrous convolution for up-sample: it is a dilated convolution which uses a dilation rate to effectively enlarge the field of view of filters without increasing the number of parameters or the amount of computation. The last improvement of the DeepLab family is DeepLabv3+ [8], which combines cascaded and parallel modules of dilated convolutions and it is the architecture used in this work.

Several other architectures have been proposed in the literature for image segmentation, including recurrent neural network based models, attention-based models and generative models. The interested reader can refer to [9] for a recent survey.

Apart from the main architecture of the network, there are a handful of other good design choices that would help achieve good performance. For example, the choice of a pretrained backbone for the encoder part of the network. Among several CNNs [10] widely used for transfer learning we tested the following models:

- MobileNet-v2 [11] is a lightweight CNN designed for mobile devices based on depthwise separable convolutions.

- ResNet18 and ResNet50 [12] are two CNNs of the ResNet family, a set of architecture based on the use of residual blocks in which intermediate layers of a block learn a residual function with reference to the block input.

- Xception [13], is a CNN architecture that relies solely on depthwise separable convolution layers.

- IncR, Inception-ResNet-v2 [14] combines the Inception architecture with residual connections. In the Inception-Resnet block, multiple sized convolutional filters are combined with residual connections, replacing the filter concatenation stage of the Inception architecture.

- EfficentNetb0 [15], is a family of CNNs designed to scale well with performance. EfficientNet-B0 is a simple mobile-size baseline architecture, the other networks of the family are obtained applying an effective compound scaling method for increasing the model size to achieve maximum accuracy gains.

In Table 1 a summary of the above models is reported.

**Table 1:** Summary of CNN models.

| Network | Depth | Size (MB) | Parameters (Millions) | Input Size |
|---|---|---|---|---|
| mobilenetv2 | 53 | 13 | 3.5 | 224×224 |
| resnet18 | 18 | 44 | 11.7 | 224×224 |
| resnet50 | 50 | 96 | 25.6 | 224×224 |
| xception | 71 | 85 | 22.9 | 299×299 |
| IncR | 164 | 209 | 55.9 | 299×299 |
| efficientnetb0 | 82 | 20 | 5.3 | 224×224 |

Also the choice of the loss function influences the way the network is trained. The most commonly used loss function for the task of image segmentation is a pixel-wise cross entropy loss. This loss treats the problem as a multi-class classification problem at pixel level comparing the class predictions to the actual label. Pixel-wise loss is calculated as the log loss summed over all the classes and averaged over all pixels. This can be a problem if some classes have unbalanced representation in the image, as training can be dominated by the most prevalent class. A possible solution is to use weighting for each class in order to counteract a class imbalance present in the dataset [2].

Another popular loss function for image segmentation is the Dice loss [16], which is based on the Sørensen-Dice similarity coefficient for measuring overlap between two segmented images. This measure ranges from 0 to 1 where a Dice coefficient of 1 denotes perfect and complete overlap. The dice loss is used in this work. Other popular loss functions for image segmentation and their usage for fast and better convergence of a model are reviewed in [17].

Moreover, the choice of the activation function can be significant. ReLU is the nonlinearity that most works use in the area, but several works have reported improved results with different activation functions [18]. In section 3 our approach for perturbing models by replacing activation layer is explained.

Finally, data augmentation can help avoid overfitting since in many applications the size of the dataset is small compared to the number of parameters in a segmentation deep neural network. We perform experiments with data augmentation, consisting in horizontal and vertical flips and rotations of 90°.

**3. Stochastic Activation Selection**

Given a neural network architecture and a pool of different activation functions, Stochastic Activation Selection consists in creating different versions of the same architecture that differ in the choice of the activation layers. This method was first introduced in [18]. The process to create a new network is based on the replacement of each activation layer (ReLU) by a new activation function which can be fixed a priori or randomly selected from the ones in the pool. This new function is substituted into the original architecture. This leads to a new network, which in the stochastic version, has different activation layers through the network. Since this is a random procedure, it yields a different network every time. Hence, we iterate the process multiple times to create many different networks that we use to create an ensemble of neural networks. We train each network independently on the same set of data and then we merge their results using the sum rule, which consists in averaging the softmax output of all the networks in the ensemble.

In our paper, we use Deeplabv3+ [8] as neural architecture. The pool of activation functions is made by ReLU and a list of its modifications proposed in the literature: ReLU [19], Leaky ReLU [20], ELU [21], PReLU [22], S-Shaped ReLU [23] (SReLU), Adaptive Piecewise Linear Unit [24] (APLU), Mexican ReLU [25] (MeLU) (with $k \in \{4,8\}$), Gaussian Linear Unit (GaLU) [18] (with $k \in \{4,8\}$), PDELU, [26], Swish (fixed and learnable) [27], Soft Root Sign [28], Mish (fixed and learnable) [29] and Soft Learnable [30].

**4. Dataset, testing protocol and metrics**

All the experiments have been carried out on the Kvasir-SEG dataset [31] which includes 1000 polyp images acquired by a high-resolution electromagnetic imaging system, with a ground-truth consisting of bounding boxes and segmentation masks. For a fair comparison with other approaches (see table 4) as [32] and [33] we use the following testing protocol: 880 images are used for training, and the remaining 120 for testing.

The image sizes vary between 332 × 487 to 1920 × 1072 pixels. For training purposes, the images are resized to the input size of each model, but for performance evaluation the predicted masks are resized back to the original dimensions (please note that other approaches evaluated performance on the resized version of the images).

We train our models with SGD optimizer for 20 epochs and a learning rate of 10e-2 (see the code for details).

Several metrics have been proposed in the literature to evaluate the performance of an image segmentation models. We report metrics for segmentation in two classes (foreground/background), which are suited to the polyp segmentation problem, anyway they can be easily extended to multiclass problems. The following metrics are most popular to quantify model accuracy:

- Accuracy / Precision / Recall / F1-score / F2-score can be defined for a bi-class problem (or for each class in case of multiclass) starting from the confusion matrix (TP, TN, FP, FN refer to the true positives, true negatives, false positives and false negatives, respectively) as follows:

$$Accuracy = \frac{TP + TN}{TP + FP + FN + TN} \quad (1)$$

$$Precision = \frac{TP}{TP + FP} \quad (2)$$

$$Recall = \frac{TP}{TP + FN} \quad (3)$$

$$F1 - score = \frac{2 \cdot TP}{2 \cdot TP + FP + FN} \quad (4)$$

$$F2 - score = \frac{5 \cdot Precision \cdot Recall}{4 \cdot Precision + Recall} \quad (5)$$

- Intersection over Union (IoU): IoU is defined as the area of intersection between the predicted segmentation map A and the ground truth map B, divided by the area of the union between the two maps:

$$IoU = \frac{|A \cap B|}{|A \cup B|} = \frac{TP}{TP + FP + FN} \quad (6)$$

- Dice: the Dice coefficient, is defined as twice the overlap area of the predicted and ground-truth maps divided by the total number of pixels. For binary maps, with foreground as the positive class, the Dice coefficient is identical to the F1-score:

$$Dice = \frac{|A \cap B|}{|A| + |B|} = \frac{2 \cdot TP}{2 \cdot TP + FP + FN} \quad (7)$$

All the above reported metrics range in [0,1] and must be maximized. The final performance is obtained averaging on the test set the performance obtained for each test image.

## 5. Experiments and discussion

The first experiment (Table 2) is aimed at comparing the different backbone networks listed in section 2. Since the size of images in Kvasir dataset is quite large we also evaluate a version of the Resnet with larger input size (i.e. 299×299 and 352×352)

**Table 2:** Experiments with different backbones.

| Backbone | IoU | Dice | F2 | Prec. | Rec. | Acc. |
|---|---|---|---|---|---|---|
| Mobilenetv2 | 0.734 | 0.823 | 0.827 | 0.863 | 0.841 | 0.947 |
| resnet18 | 0.759 | 0.844 | 0.845 | 0.882 | 0.856 | 0.952 |
| resnet50 | 0.751 | 0.837 | 0.836 | 0.883 | 0.845 | 0.952 |
| xception | 0.699 | 0.799 | 0.792 | 0.870 | 0.800 | 0.943 |
| IncR | 0.793 | 0.871 | 0.878 | 0.889 | 0.892 | 0.961 |
| efficientnetb0 | 0.705 | 0.800 | 0.801 | 0.860 | 0.814 | 0.944 |
| resnet18-299 | 0.782 | 0.863 | 0.870 | 0.881 | 0.883 | 0.959 |
| resnet50-299 | 0.798 | 0.872 | 0.876 | **0.898** | 0.886 | 0.962 |
| resnet18-352 | 0.787 | 0.865 | 0.871 | 0.891 | 0.884 | 0.960 |
| resnet50-352 | **0.801** | **0.872** | **0.884** | 0.881 | **0.900** | **0.964** |

The second experiment (Table 3) is aimed at designing effective ensembles by varying the activation functions. Each ensemble is fusion by the sum rule of 14 models (since we use 14 activation functions). The ensemble name is the concatenation of the name of the backbone network and a string to identify the creation approach:

- act: each network is obtained by deterministically substituting each activation layer by one of the activation functions of section 3 (the same function for all the layers, but a different function for each network)
- sto: ensembles of stochastic models, whose activation layers have been replaced by a randomly selected activation function (which may be different for each layer)
- relu: an ensemble of original models which differ only for the random initialization before training. It means that all the starting models in the ensemble are the same, except for the initialization.

**Table 3:** Experiments on ensembles.

| Ensemble name – 14 models | IoU | Dice | F2 | Prec. | Rec. | Acc. |
|---|---|---|---|---|---|---|
| resnet18_act | 0.774 | 0.856 | 0.856 | 0.888 | 0.867 | 0.955 |
| resnet18_relu | 0.774 | 0.858 | 0.858 | 0.892 | 0.867 | 0.955 |
| resnet18_sto | 0.780 | 0.860 | 0.857 | 0.898 | 0.864 | 0.956 |
| resnet50_act | 0.779 | 0.858 | 0.859 | 0.894 | 0.869 | 0.957 |
| resnet50_relu | 0.772 | 0.855 | 0.858 | 0.889 | 0.870 | 0.955 |
| resnet50_sto | 0.779 | 0.859 | 0.864 | 0.891 | 0.877 | 0.957 |
| IncR_sto | 0.804 | 0.877 | 0.881 | 0.903 | 0.893 | 0.964 |
| resnet50-352_sto | **0.820** | **0.885** | **0.888** | **0.915** | **0.896** | **0.966** |

Clearly using larger input sizes boost the performance of Resnet50 (see Table 2); the best performance, among the ensembles, is obtained by resnet50-352_sto.

Finally, in Table 4 a comparison with some state-of-the-art results is reported.

**Table 4:** State-of-the-art approaches using exactly the same testing protocol (all values are those reported in the reference paper, except for # our approaches). The results of many methods are reported in [32], please read it for the original reference of a given approach.

| Method | IoU | Dice | F2 | Prec. | Rec. | Acc. |
|---|---|---|---|---|---|---|
| resnet50-352 # | 0.801 | 0.872 | 0.884 | 0.881 | 0.900 | 0.964 |
| resnet50-352_sto # | 0.820 | 0.885 | 0.888 | 0.915 | 0.896 | 0.966 |
| U-Net [32] | 0.471 | 0.597 | 0.598 | 0.672 | 0.617 | 0.894 |
| ResUNet [32] | 0.572 | 0.69 | 0.699 | 0.745 | 0.725 | 0.917 |
| ResUNet++ [32] | 0.613 | 0.714 | 0.72 | 0.784 | 0.742 | 0.917 |
| FCN8 [32] | 0.737 | 0.831 | 0.825 | 0.882 | 0.835 | 0.952 |
| HRNet [32] | 0.759 | 0.845 | 0.847 | 0.878 | 0.859 | 0.952 |
| DoubleUNet [32] | 0.733 | 0.813 | 0.82 | 0.861 | 0.84 | 0.949 |
| PSPNet [32] | 0.744 | 0.841 | 0.831 | 0.89 | 0.836 | 0.953 |
| DeepLabv3+ ResNet50 [32] | 0.776 | 0.857 | 0.855 | 0.891 | 0.861 | 0.961 |
| DeepLabv3+ ResNet101 [32] | 0.786 | 0.864 | 0.857 | 0.906 | 0.859 | 0.961 |
| U-Net ResNet34 [32] | 0.81 | 0.876 | 0.862 | 0.944 | 0.86 | 0.968 |
| DDANet [34] | 0.78 | 0.858 | --- | 0.864 | 0.888 | --- |
| HarDNet-MSEG [33] | 0.848 | 0.904 | 0.915 | 0.907 | 0.923 | 0.969 |
| ColonSegNet [35] | --- | 0.821 | 0.821 | 0.843 | 0.850 | 0.949 |

Our best approach obtains best performance but HarDNet-MSEG (original paper), notice that our approach strongly outperforms several other deep learning approaches.

## 6. Conclusions

Semantic segmentation is a very important topic in medical-image analysis. In this paper our aim is to optimize the performance in polyp segmentation during colonoscopy examinations.

We have compared several convolution neural network architectures, including ResNet, Xception, EfficentNet, MobileNet and different methods for building ensemble of CNN.

Our reported results show that the best ensemble obtains state of the art performance in the tested dataset (Kvasir-SEG dataset). To reproduce our results the MATLAB source code is available at GitHub: https://github.com/LorisNanni.

**Acknowledgment**

We gratefully acknowledge the support of NVIDIA Corporation for the "NVIDIA Hardware Donation Grant" of a Titan X used in this research.

**References**

[1] L. Roncucci, F. Mariani, Prevention of colorectal cancer: How many tools do we have in our basket?, Eur. J. Intern. Med. 26 (2015) 752–756. https://doi.org/10.1016/j.ejim.2015.08.019.

[2] E. Shelhamer, J. Long, T. Darrell, Fully Convolutional Networks for Semantic Segmentation, IEEE Trans. Pattern Anal. Mach. Intell. (2017). https://doi.org/10.1109/TPAMI.2016.2572683.

[3] H. Noh, S. Hong, B. Han, Learning deconvolution network for semantic segmentation, in: Proc. IEEE Int. Conf. Comput. Vis., 2015. https://doi.org/10.1109/ICCV.2015.178.

[4] O. Ronneberger, P. Fischer, T. Brox, U-net: Convolutional networks for biomedical image segmentation, in: Lect. Notes Comput. Sci. (Including Subser. Lect. Notes Artif. Intell. Lect. Notes Bioinformatics), 2015. https://doi.org/10.1007/978-3-319-24574-4_28.

[5] V. Badrinarayanan, A. Kendall, R. Cipolla, SegNet: A Deep Convolutional Encoder-Decoder Architecture for Image Segmentation, IEEE Trans. Pattern Anal. Mach. Intell. 39 (2017) 2481–2495.

[6] K. Simonyan, A. Zisserman, Very Deep Convolutional Networks for Large-Scale Image Recognition, CoRR. abs/1409.1 (2015).

[7] L.C. Chen, G. Papandreou, I. Kokkinos, K. Murphy, A.L. Yuille, DeepLab: Semantic Image Segmentation with Deep Convolutional Nets, Atrous Convolution, and Fully Connected CRFs, IEEE Trans. Pattern Anal. Mach. Intell. (2018). https://doi.org/10.1109/TPAMI.2017.2699184.

[8] L.C. Chen, Y. Zhu, G. Papandreou, F. Schroff, H. Adam, Encoder-decoder with atrous separable convolution for semantic image segmentation, in: Lect. Notes Comput. Sci. (Including Subser. Lect. Notes Artif. Intell. Lect. Notes Bioinformatics), 2018. https://doi.org/10.1007/978-3-030-01234-2_49.

[9] S. Minaee, Y. Boykov, F. Porikli, A. Plaza, N. Kehtarnavaz, D. Terzopoulos, Image segmentation using deep learning: A Survey, ArXiv. (2020).

[10] A. Khan, A. Sohail, U. Zahoora, A.S. Qureshi, A survey of the recent architectures of deep convolutional neural networks, Artif. Intell. Rev. (2020). https://doi.org/10.1007/s10462-020-09825-6.

[11] M. Sandler, A. Howard, M. Zhu, A. Zhmoginov, L.-C. Chen, Mobilenetv2: Inverted residuals and linear bottlenecks, in: Proc. IEEE Conf. Comput. Vis. Pattern Recognit., 2018: pp. 4510–4520.

[12] K. He, X. Zhang, S. Ren, J. Sun, Deep Residual Learning for Image Recognition, 2016 IEEE Conf. Comput. Vis. Pattern Recognit. (2016) 770–778.

[13] F. Chollet, Xception: Deep learning with depthwise separable convolutions, in: Proc. - 30th IEEE Conf. Comput. Vis. Pattern Recognition, CVPR 2017, 2017. https://doi.org/10.1109/CVPR.2017.195.

[14] C. Szegedy et al., Inception-v4, Inception-ResNet and the Impact of Residual Connections on Learning., Proc. IEEE Int. Conf. Comput. Vis. 115 (2017) 4278–4284. https://doi.org/10.1145/3038912.3052569.

[15] M. Tan, Q. V. Le, EfficientNet: Rethinking model scaling for convolutional neural networks, in: 36th Int.

Conf. Mach. Learn. ICML 2019, 2019.

[16] C.H. Sudre, W. Li, T. Vercauteren, S. Ourselin, M. Jorge Cardoso, Generalised dice overlap as a deep learning loss function for highly unbalanced segmentations, in: Lect. Notes Comput. Sci. (Including Subser. Lect. Notes Artif. Intell. Lect. Notes Bioinformatics), 2017. https://doi.org/10.1007/978-3-319-67558-9_28.

[17] S. Jadon, A survey of loss functions for semantic segmentation, in: 2020 IEEE Conf. Comput. Intell. Bioinforma. Comput. Biol. CIBCB 2020, 2020. https://doi.org/10.1109/CIBCB48159.2020.9277638.

[18] L. Nanni, A. Lumini, S. Ghidoni, G. Maguolo, Stochastic selection of activation layers for convolutional neural networks, Sensors (Switzerland). (2020). https://doi.org/10.3390/s20061626.

[19] X. Glorot, A. Bordes, Y. Bengio, Deep Sparse Rectifier Neural Networks, in: AISTATS, 2011.

[20] A.L. Maas, A.Y. Hannun, A.Y. Ng, Rectifier nonlinearities improve neural network acoustic models, ICML. 28 (2013) 6. https://doi.org/10.1016/0010-0277(84)90022-2.

[21] D.A. Clevert, T. Unterthiner, S. Hochreiter, Fast and accurate deep network learning by exponential linear units (ELUs), in: 4th Int. Conf. Learn. Represent. ICLR 2016 - Conf. Track Proc., 2016.

[22] K. He, X. Zhang, S. Ren, J. Sun, Delving deep into rectifiers: Surpassing human-level performance on imagenet classification, in: Proc. IEEE Int. Conf. Comput. Vis., 2015: pp. 1026–1034.

[23] X. Jin, C. Xu, J. Feng, Y. Wei, J. Xiong, S. Yan, Deep Learning with S-shaped Rectified Linear Activation Units, in: AAAI, 2016.

[24] F. Agostinelli, M. Hoffman, P. Sadowski, P. Baldi, Learning activation functions to improve deep neural networks, in: 3rd Int. Conf. Learn. Represent. ICLR 2015 - Work. Track Proc., 2015.

[25] G. Maguolo, L. Nanni, S. Ghidoni, Ensemble of Convolutional Neural Networks Trained with Different Activation Functions, CoRR. abs/1905.0 (2019).

[26] Q. Cheng, H. Li, Q. Wu, L. Ma, N.N. King, Parametric Deformable Exponential Linear Units for deep neural networks, Neural Networks. (2020).

[27] P. Ramachandran, B. Zoph, Q. V Le, Searching for Activation Functions, CoRR. abs/1710.0 (2017).

[28] Y. Zhou, D. Li, S. Huo, S.-Y. Kung, Soft-Root-Sign Activation Function, ArXiv Prepr. ArXiv2003.00547. (2020).

[29] D. Misra, Mish: A self regularized non-monotonic neural activation function, ArXiv Prepr. ArXiv1908.08681. (2019).

[30] L. Nanni, A. Lumini, S. Ghidoni, G. Maguolo, Comparisons among different stochastic selection of activation layers for convolutional neural networks for healthcare, ArXiv Prepr. ArXiv2011.11834. (2020).

[31] D. Jha, P.H. Smedsrud, M.A. Riegler, P. Halvorsen, T. de Lange, D. Johansen, H.D. Johansen, Kvasir-SEG: A Segmented Polyp Dataset, in: Lect. Notes Comput. Sci. (Including Subser. Lect. Notes Artif. Intell. Lect. Notes Bioinformatics), 2020. https://doi.org/10.1007/978-3-030-37734-2_37.

[32] D. Jha, S. Ali, H.D. Johansen, D. Johansen, J. Rittscher, M.A. Riegler, P. Halvorsen, Real-time polyp detection, localisation and segmentation in colonoscopy using deep learning, ArXiv. (2020).

[33] C.-H. Huang, H.-Y. Wu, Y.-L. Lin, HarDNet-MSEG: A Simple Encoder-Decoder Polyp Segmentation Neural Network that Achieves over 0.9 Mean Dice and 86 FPS, (2021).

[34] Nikhil Kumar Tomar, Debesh Jha, Sharib Ali, Håvard D. Johansen, Dag Johansen, Michael A. Riegler and Pål Halvorsen, DDANet: Dual Decoder Attention Network for Automatic Polyp Segmentation, https://arxiv.org/pdf/2012.15245v1.pdf

[35] Debesh Jha, Sharib Ali, Nikhil Kumar Tomar, Håvard D. Johansen, Dag Johansen, Jens Rittscher, Michael A. Riegler, Pål Halvorsen, Real-Time Polyp Detection, Localization and Segmentation in Colonoscopy Using Deep Learning, https://arxiv.org/pdf/2011.07631.pdf